%% file: root.tex
\documentclass[letterpaper, 10 pt, conference]{ieeeconf}

\IEEEoverridecommandlockouts
\usepackage{afterpage}
\usepackage{xspace}
\usepackage{amssymb,paralist,epsfig,standalone,bm,placeins}  
\usepackage{graphicx} 
\usepackage{xcolor}
\usepackage{multirow}
\usepackage{makecell}
\usepackage{multicol}
\usepackage[utf8]{inputenc}
\usepackage{amsmath,amsfonts,booktabs,cite} 
\usepackage{siunitx,textcomp} 
\usepackage[hidelinks]{hyperref} 
\let\labelindent\relax
\usepackage[inline]{enumitem} 
\usepackage[nolist]{acronym} 
\usepackage{caption}
\usepackage{tikz,standalone,pgfplots}
\pgfplotsset{width=10cm,compat=1.9}
\usepackage{siunitx}
\usetikzlibrary{patterns}
\usepackage{pgfplots, pgfplotstable}
\graphicspath{{figures_tex/}}

\usepackage{changes}
\usepackage[ruled,vlined]{algorithm2e}
\usepackage{subcaption}

\makeatletter
\newcommand{\removelatexerror}{\let\@latex@error\@gobble}
\let\@oldmaketitle\@maketitle
\renewcommand{\@maketitle}{\@oldmaketitle
    \centering
        \centering
        \includegraphics[width=\linewidth]{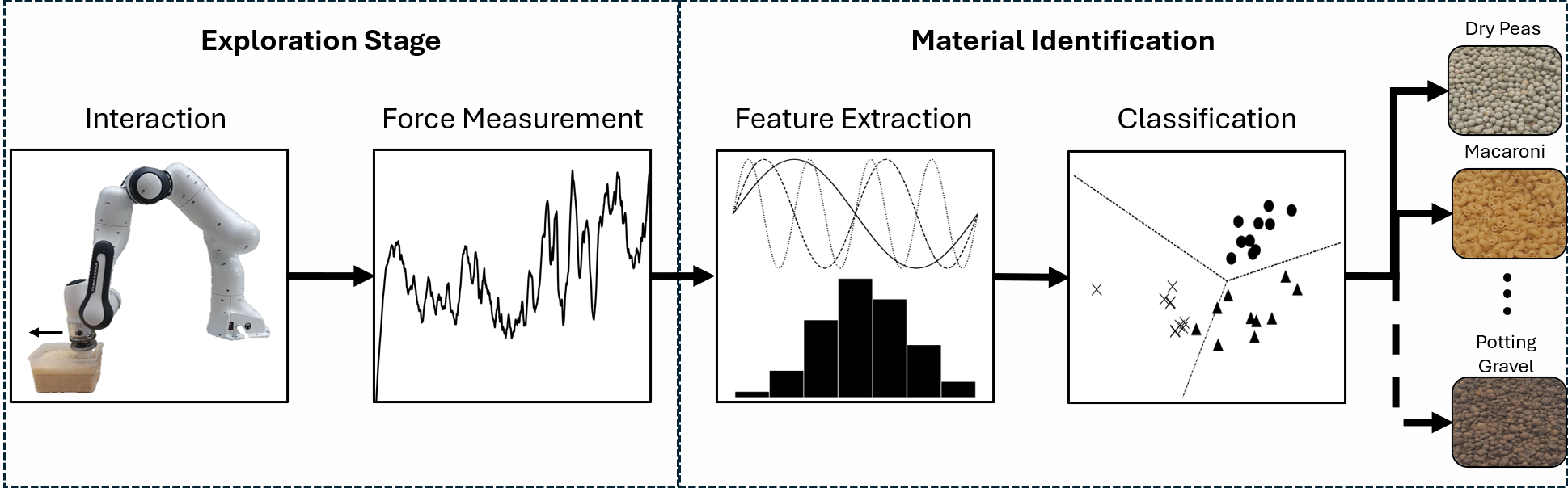}
  \captionof{figure}{To identify granular materials, we propose to interactively explore the material, recording interaction forces. The raw force measurements are then processed to identify meaningful features to classify the material. 
 \label{fig:fig1} 
}}
\makeatother
\pgfplotsset{compat=1.3}
\input{acronyms.tex}

\input{macro.tex}

\definecolor{findOptimalPartition}{HTML}{D7191C}
\definecolor{storeClusterComponent}{HTML}{FDAE61}
\definecolor{dbscan}{HTML}{ABDDA4}
\definecolor{constructCluster}{HTML}{2B83BA}

\title{\LARGE \bf Interactive Identification of Granular Materials \\ using Force Measurements}

\author{Samuli~Hynninen$^{1}$, Tran~Nguyen~Le$^{2}$, Ville~Kyrki$^{1}$%
\thanks{This work was financially supported by Business Finland (decision 9249/31/2021).} \thanks{$^{1}$ Intelligent Robotics Group at the Department of Electrical Engineering and
Automation, School of Electrical Engineering, Aalto University, Finland.
\texttt{\{firstname.lastname\}{@}aalto.fi}}
\thanks{$^{2}$ Section of Mechanical Technology at the Department of Engineering Technology and Didactics, Technical University of Denmark, Denmark. Part of the research presented in this work was conducted when Tran Nguyen Le was at Aalto University. \texttt{tngle@dtu.dk}}
}

\begin{document}
\maketitle
\thispagestyle{empty}
\pagestyle{empty}


\begin{abstract}

%

Despite the potential the ability to identify granular materials creates for applications such as robotic cooking or earthmoving, granular material identification remains a challenging area, existing methods mostly relying on shaking the materials in closed containers. This work presents an interactive material identification framework that enables robots to identify a wide range of granular materials using only force-torque measurements. Unlike prior works, the proposed approach uses direct interaction with the materials. The approach is evaluated through experiments with a real-world dataset comprising 11 granular materials, which we also make publicly available. Results show that our method can identify a wide range of granular materials with near-perfect accuracy while relying solely on force measurements obtained from direct interaction. Further, our comprehensive data analysis and experiments show that a high-performancefeature space must combine features related to the force signal's time-domain dynamics and frequency spectrum. We account for this by proposing a combination of the raw signal and its high-frequency magnitude histogram as the suggested feature space representation. We show that the proposed feature space outperforms baselines by a significant margin. The code and data set are available at: \url{https://irobotics.aalto.fi/identify_granular/}.

\end{abstract}

\input{sections/introduction}

\input{sections/related_work}

\input{sections/method}

\input{sections/data_analysis_sep24}

\input{sections/experiment}

\input{sections/limitations_future}
\input{sections/conclusion}



\bibliographystyle{IEEEtran}
\bibliography{refs}

\end{document}

%% file: acronyms.tex
\newacro{ftsensor}[F/T sensor]{force-torque sensor}
\newacro{ecoc}[ECOC]{error-correcting output codes}
\newacro{svm}[SVM]{support vector machines}

\newacro{pbd}[PBD]{Position-based Dynamics}
\newacro{fem}[FEM]{Finite Element Method}
\newacro{dnn}[DNN]{Deep Neural Network}
\newacro{fcn}[FCN]{fully-convolutional network}

%% file: macro.tex
\newcommand{\figref}[1]{\hyperref[#1]{Fig.~\ref*{#1}}}
\newcommand{\tabref}[1]{\hyperref[#1]{Table~\ref*{#1}}}
\newcommand{\secref}[1]{\hyperref[#1]{Section~\ref*{#1}}}
\newcommand{\algoref}[1]{\hyperref[#1]{Algorithm~\ref*{#1}}}

\newcommand{\ra}[1]{\renewcommand{\arraystretch}{#1}}
\newcommand{\tbs}[1]{\renewcommand{\tabcolsep}{#1pt}}

%% file: sections/introduction.tex
\section{Introduction}
\label{sec:introduction}
The ability to accurately identify granular materials holds great potential across various domains of robotics, paving the way for future advancements in assistive robotics as well as autonomous working machines. In assistive robotics, the ability to distinguish different ingredients in food from each other is valuable to perform tasks such as assistive feeding or cooking assistance. Similarly, in industries such as mining or construction, autonomous work machines benefit from being able to differentiate between granular materials for tasks such as earthmoving or soil cultivation. 

Most of the previous work on granular material identification or parameter estimation is done outside the robotic manipulation community, mostly in terramechanics and geoengineering, often utilizing tools such as Bevameter \cite{apfelbeck2011systematic, kruger2023experimental} or cone penetrometer \cite{white2022cpt, bian2025assessment}. Consequently, the approaches require specialized hardware with limited manipulability. This significantly limits the methods' applicability in various domains of robotics and autonomous machines, where agility and multipurpose tools and sensors are often valued. In the robotic manipulation context, existing works mostly rely on tactile array sensors or audio sensing and identify materials based on their vibration signature when the material is shaken in a closed container \cite{eppe2018deep, jin2019open,chen2016learning, wang2023stev}. However, in tasks such as earthmoving, the materials are not in closed containers, which limits the suitability of the methods based on vibration signatures.  

To address this gap, we introduce a novel interactive perception framework that identifies granular materials by interacting with them directly. Specifically, our method measures feedback from the interaction with a standard 6-axis \ac{ftsensor} mounted on the robot's end effector, which is partially submerged in the material during a predefined motion. The collected F/T signals are subsequently classified based on the features extracted from the raw F/T measurements. We use custom-designed features based on our qualitative data analysis findings.  

To the best of the authors' knowledge, this work is the first attempt to identify granular materials using only a regular \ac{ftsensor} for perception. 
The contributions of this work include:
\begin{itemize}
    \item  A novel interactive perception framework, enabling accurate identification of various granular materials, improving robots' ability to understand and respond to the environment.
    \item A novel real-world dataset of force-torque measurement sequences containing a total of 682 samples from 11 materials.
    \item A comprehensive analysis of the dataset, facilitating a deeper understanding of its nature, the distinguishability of materials, and identification of their distinctive features.
\end{itemize}

%% file: sections/related_work.tex
\section{Related work}
\label{sec:related_work}
To put our work in context, we next review two complementary viewpoints, interactive perception, and granular material identification. Considering interactive perception, we position our focus on works that rely on 6-axis F/T sensors since they are closest related to our work. 
\subsection{Interactive perception with 6-axis F/T sensor}

The majority of works on interactive perception with F/T sensors focus on identifying surface materials~\cite{liu12} or estimating surface properties~\cite{le2021probabilistic} of rigid bodies. Typically, these studies involve attaching the F/T sensor to an artificial fingertip or similar end-effector, which interacts with the material surface to explore its properties. For example, ~\cite{lam2014study} uses this approach to compare different classifiers, such as k-nearest neighbors method, support vector machines, and neural networks on their ability to identify 18 different surface materials. Similarly, ~\cite{markert2021fingertip} proposed an interaction procedure for an artificial fingertip, which first gathers F/T data from 21 different surface materials, then classifies the materials with high accuracy using features derived from the F/T measurements' spectral distributions. Furthermore, ~\cite{le2021probabilistic} integrated F/T sensing with visual information to estimate friction maps for multi-material rigid objects.

In addition, F/T sensors have also been used for object recognition and property estimation for rigid objects. For example, ~\cite{higy2016combining} used 6-DoF F/T sensors combined with robot joint positions to perform haptic classification of rigid objects. In~\cite{pavlic2023robotscale}, the authors proposed a method that employs a 6-axis F/T sensor attached to the robot's two-fingered end effector to estimate the mass and inertia of rigid objects.

Prior research highlights that F/T sensors can provide valuable information for various robotic perception tasks. Our work expands the scope of F/T sensing by introducing a novel application area,  granular materials, and specifically their identification.

\subsection{Granular Material Identification}
Alongside with liquid identification and property estimation \cite{saal2010active, matl2019haptic, zhu2024liquids, elbrechter2015discriminating}, research on identification of granular materials is sparse in the robotics literature. The problem is better studied in terramechanics and geoengineering, where a number of approaches have been proposed to characterize and identify different types of soil \cite{apfelbeck2011systematic, kruger2023experimental, white2022cpt, bian2025assessment}. However, these methods rely on specialized and possibly large hardware, making them inapplicable for many robotic manipulation tasks. For example, the robot may have limited capacity to carry tools, making multipurpose tools and sensors a preferred option, or the task and the environment may require agile maneuvering that is not possible with physically large tools. 

In robotics, the existing works on granular material identification have utilized audio recordings~\cite{eppe2018deep, jin2019open}, vibration measurements~\cite{chen2016learning, wang2023stev}, and capacitance sensing~\cite{hu2024robocaproboticclassificationprecision}. In addition to identification, parameter estimation of granular materials has been studied \cite{matl2020inferring, guo2023estimating}. Furthermore, tactile sensor have been proposed for estimating the amount of granular materials and fluids in deformable containers~\cite{liu2023content}.

In audio-based identification methods, a robot positions a closed container close to one or more microphones, shakes the container, and classifies the material based on the audio patterns created by colliding granules inside the container. For instance, ~\cite{eppe2018deep} used a humanoid robot to shake a closed capsule and used a neural network to classify the collected audio samples. Similarly, ~\cite{jin2019open} used a robot arm to rotate capsules filled with granular materials, recorded audio samples, pre-processed the samples by a noise reduction algorithm, and transformed them into the Mel-Frequency Cepstral Coefficient space to perform classification. 

A major drawback with the audio-based methods is their proneness to get disturbed by background noise, as demonstrated in~\cite{eppe2018deep}. This might limit the approaches' applicability in loud environments such as industrial production or construction sites. Furthermore, limited identification performance is reported for materials that do not make loud sounds when colliding with the container. 

Vibration-based methods complement microphone data with other sensor modalities or use specific vibration sensors. The method in~\cite{chen2016learning} is similar to audio-based methods since the work utilizes recordings from two microphones and a 3-axis accelerometer and classifies the recorded signals from a shaking action. In \cite{wang2023stev}, a novel electronic skin vibration sensor was introduced and applied successfully to recognize granular materials inside a closed bottle. A significant disadvantage of the vibration sensing-based methods is that they require either specialized hardware or a complicated sensor setup, which may be expensive and difficult to set up in dynamic real-world scenarios.

From a general viewpoint, we argue that the shaking-a-closed-container interaction is suboptimal when the classification information is used for manipulation-related downstream tasks. This is because indirect interaction barely gives information about the material's friction and flowing resistance properties, which are highly important for planning downstream manipulation actions such as scooping or transportation. Further, in tasks such as earthmoving, the materials are not in closed containers, and thus the shaking approach is not directly applicable.

To summarize, our work aims to fill several gaps in the existing literature by proposing a granular material identification method that is not prone to background noise, can easily deal with materials that do not produce a loud sound when colliding, uses only a single, multipurpose sensing modality, thus requiring no specialized or complex hardware setups, and operates in the way that the interaction itself gives directly information about the dynamic properties of the material.

%% file: sections/method.tex
\section{Interactive Material Identification}
\label{sec:method}
\begin{figure*}
    \centering
    \begin{subfigure}[!ht]{0.4\textwidth}
        \centering
        \includegraphics[width=0.9\linewidth]{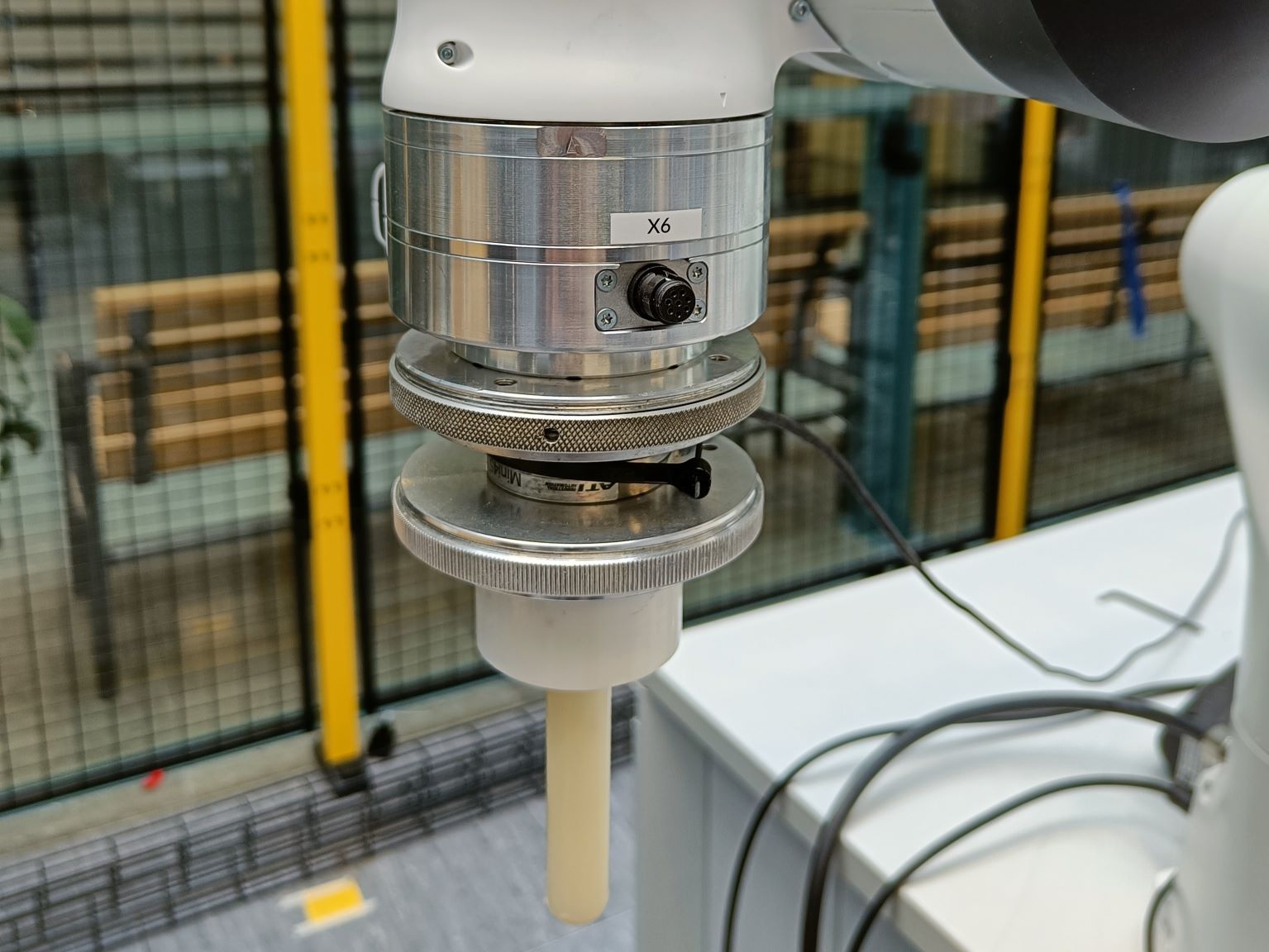}
        \captionsetup{justification=centering}
        \caption{F/T sensor and end-effector}
        \label{fig:end_effector}
    \end{subfigure}
    \hfill
    \begin{subfigure}[!ht]{0.4\textwidth}
        \centering
        \includegraphics[width=0.9\linewidth]{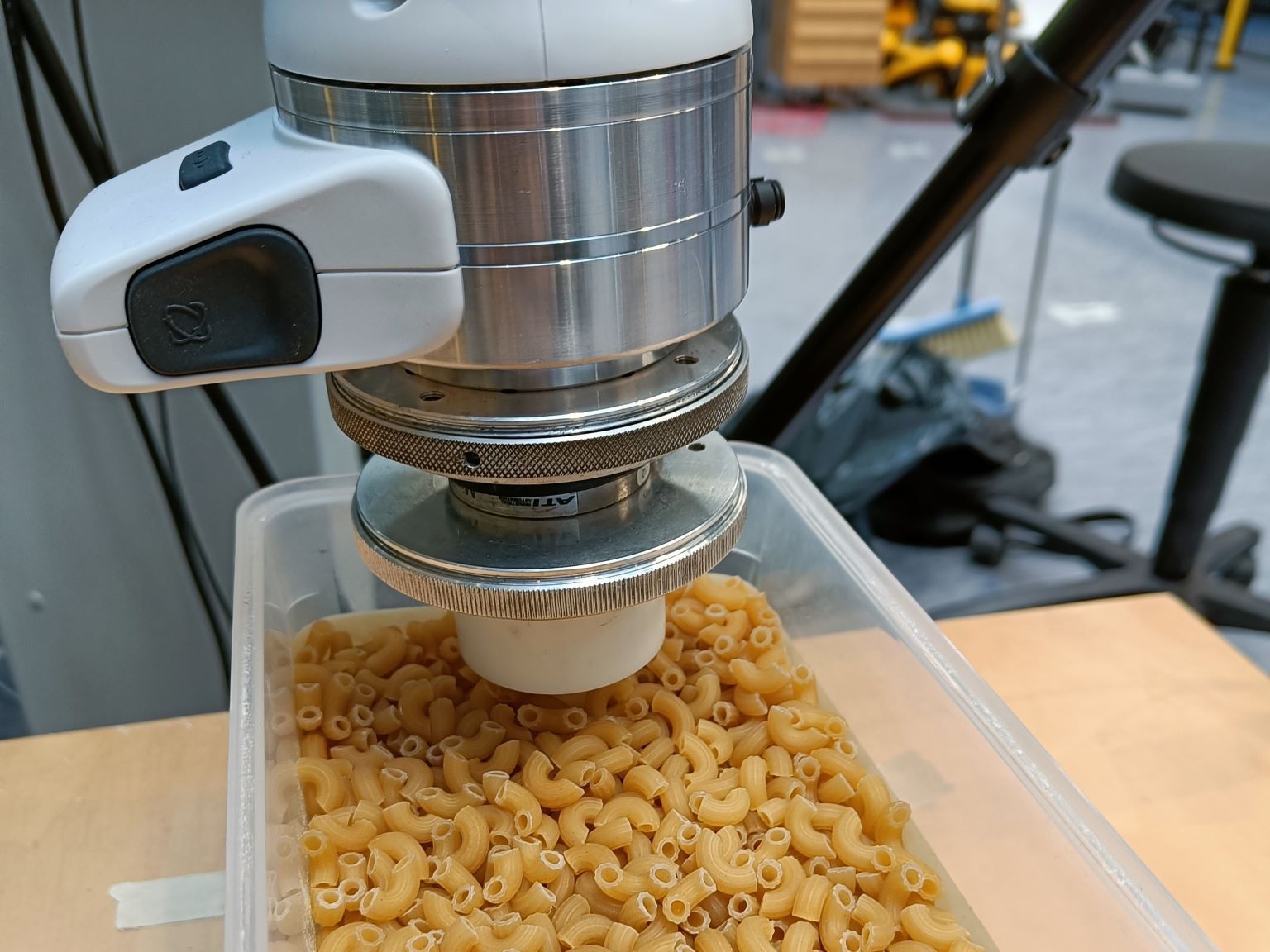}
        \captionsetup{justification=centering}
        \caption{Starting position}
        \label{fig:starting_position}
    \end{subfigure}
    \hfill
    \caption{Hardware setup and starting position}
\end{figure*}
The proposed interactive material identification framework consists of three parts, as depicted in \ref{fig:fig1}: exploration, feature extraction, and classification. At the exploration stage, the robot interacts with the materials and collects measurements with an F/T sensor. The feature extraction stage is used to process the raw F/T measurements to enhance their classifiability. Finally, we employ a classification model based on \ac{ecoc} and \ac{svm} to identify the materials. 

\subsection{Exploration}
The exploration stage involves the robot interacting with materials directly in an open container. The robot interacts with the materials using a cylinder-shaped end effector, whose height is 80 mm and diameter 17 mm. The six-axis F/T sensor is attached between the robot's wrist and the end-effector, as depicted in Fig.~\ref{fig:end_effector}. In this work, we utilize a 7-DoF Franka Emika Panda robot arm and an ATI Mini45 F/T sensor. The F/T sensor measures data at a sampling frequency of \SI{500}{\hertz} throughout the motion. 

To maintain a standardized exploration procedure, the height of the material surface and the robot starting position are predetermined and maintained consistently. The robot initiates its interaction with the material from the designated starting position, ensuring that the end effector is nearly fully submerged in the material, as depicted in Figure~\ref{fig:starting_position}. We employ a Cartesian velocity controller to guarantee constant velocity regardless of the material. The exploration trajectory entails linear motion along the x-axis at a velocity of \SI{0.03}{\meter\per\second} for $3.2$ seconds. We decided to use a straight, one-way motion trajectory for simplicity, and our results show the simple approach is sufficient to capture the materials' distinguishing characteristics. The choice on the velocity and duration of the motion was made based on our expectation on what is required to bring forth varying dynamic properties of different materials. The recording of the F/T data is started a fraction of a second before the robot begins the exploration trajectory to ensure that the information-rich beginning of the interaction is fully recorded. 

\subsection{Feature Extraction}
To enhance the classifiability of the materials, we designed a feature space that considers their distinguishing characteristics, which are further discussed in Section~\ref{sec:dataset}. Specifically, we use the raw signal augmented by its High-Frequency Magnitude Histogram (HFMH), computed separately for each axis of the input F/T signal. In short, we define HFMH as the magnitude histogram of the F/T signal after high-pass filtering. We use a histogram with 100 bins evenly spaced in the range $[-1.5, 1.5]$. The range was defined to correspond to the maximum and minimum magnitudes present in the high-pass filtered F/T signals. For high-pass filtering, we use the 8th orded Butterworth filter ~\cite{proakisdigital}, designed with respect to the cutoff frequency of \SI{23}{\hertz}. 
 This cutoff frequency was found to be optimal by trial and error. All in all, the feature extraction stage results in 10200-dimensional feature vectors from which the raw time domain signal takes $9600$ and the HFMH representation $600$ dimensions. 
\subsection{Classification}
We use the trinary \ac{ecoc} model~\cite{escalera2008decoding} for our multiclass classification task. The ECOC model divides the multiclass classification problem into a set of binary classification tasks and includes a coding and decoding scheme to determine the final class for the input. For a $K$-class classification task, we have $L = K\cdot(K-1)/2$ binary learners. Formally, the ECOC decision rule is then defined as
\begin{equation}
    \hat{y} = \underset{k \in \{1, \ldots, K\}}{\mathrm{arg\,min}}\; \dfrac{\sum_{l=1}^L |m_{kl}|g(m_{kl}, s_l)}{\sum_{l=1}^L |m_{kl}|},
\end{equation}
where $\hat{y}$ is the predicted class, $m_{kl}$ is an element of the 
 trinary coding matrix $M \in \{-1, 0, +1\}^{K\times L}$, and
 \begin{equation}
     g(m_{kl}, s_l) = \max\{0, 1 - m_{kl}s_l\} / 2
 \end{equation}
 is the binary loss. Variable $s_l$ stands for the binary classifier score for the input sample. For a linear kernel SVM, $s_l$ is given by
 \begin{equation}
     s_l = \mathbf{w}_l^\top\mathbf{x} + b_l,
 \end{equation}
 where $\mathbf{x}$ is the input feature vector, and $\mathbf{w}_l$ is the weight vector given by the solution of the optimization problem
\begin{equation}
    \begin{aligned}
        & \min_{\mathbf{w}_l, b} \quad \frac{1}{2} \|\mathbf{w}_l\|^2 \\
        & \text{s.t.} \quad y_i (\mathbf{w}_l^\top \mathbf{x}_i - b) \geq 1 \quad \forall i \in \{1, \ldots, n\}
    \end{aligned}
\end{equation}
where $y_i \in \{-1, 1\}$ is the class label of $i$-th sample.
Linear kernels were experimentally found to be most suitable for our task. They also allow a fair comparison when we compare the proposed feature space representation to baseline feature spaces.

The decision to use ECOC and SVMs with hand-designed feature extraction instead of an end-to-end data-driven approach, such as a classification neural network, was driven by two main reasons. Firstly, the time-intensive nature of real-world data collection makes it too costly to collect a sufficient dataset to train a data-hungry neural network. Secondly, transparent manually engineered features provide explainability and enhance the understanding of the problem, materials, and measured F/T signals themselves, laying the groundwork for further advancements in the field. Moreover, the explainability may facilitate the integration of the classification model into downstream planning and control algorithms.

%% file: sections/data_analysis_sep24.tex
\section{Dataset and qualitative data analysis}
\label{sec:dataset}
\subsection{Dataset collection}
We employed the exploration procedure described in Section \ref{sec:method} to collect a dataset of 62 individual samples from each of the 11 materials, totaling 682 samples. We do not split these into fixed training and test sets, but instead we use repeated random partitions as we describe in \ref{sec:exp_and_res}. These materials span a range of everyday items, including food ingredients, gardening materials, and animal care supplies. The entire set of the materials is shown in Fig.~\ref{fig:data_kuvat}. The materials were chosen to reflect diversity in density, grain size, grain shape, and surface properties of the grains. However, also some similar materials, such as gardening clay granules and cat litter (which is also a type of clay granule), were selected to increase the difficulty of the identification task and to determine the bounds of the proposed method.

The materials were initially poured into a plastic container, while accurately measuring the desired amount by markers in the container. The material surface was smoothed before the first measurement and every subsequent iteration. To prevent material compression during the data collection, the material was carefully stirred in the container after each measurement iteration. Furthermore, fresh material was introduced during the data collection for materials susceptible to grain breakage from repeated interaction with the end effector (such as potting gravel, clay granules, cat litter, and oat flakes).

\subsection{Qualitative data analysis}
\label{sec:data_analysis_B}
\begin{figure*}[t]
    \centering
    \begin{subfigure}[t]{0.32\linewidth}
        \centering
        \includegraphics[width=\linewidth]{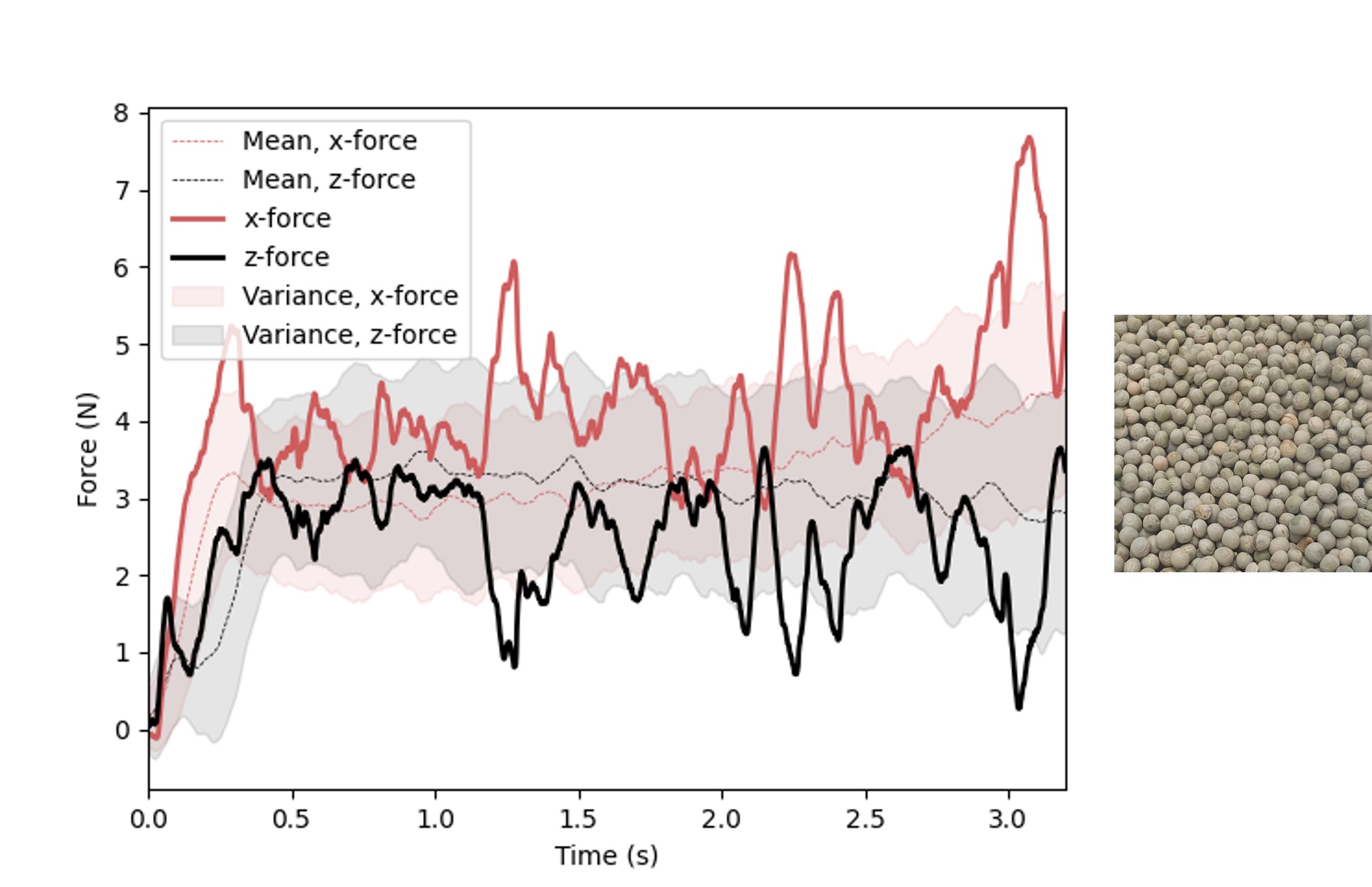}
        \captionsetup{justification=centering}
        \caption{Dry peas}
    \end{subfigure}
    \begin{subfigure}[t]{0.32\linewidth}
        \centering
        \includegraphics[width=\linewidth]{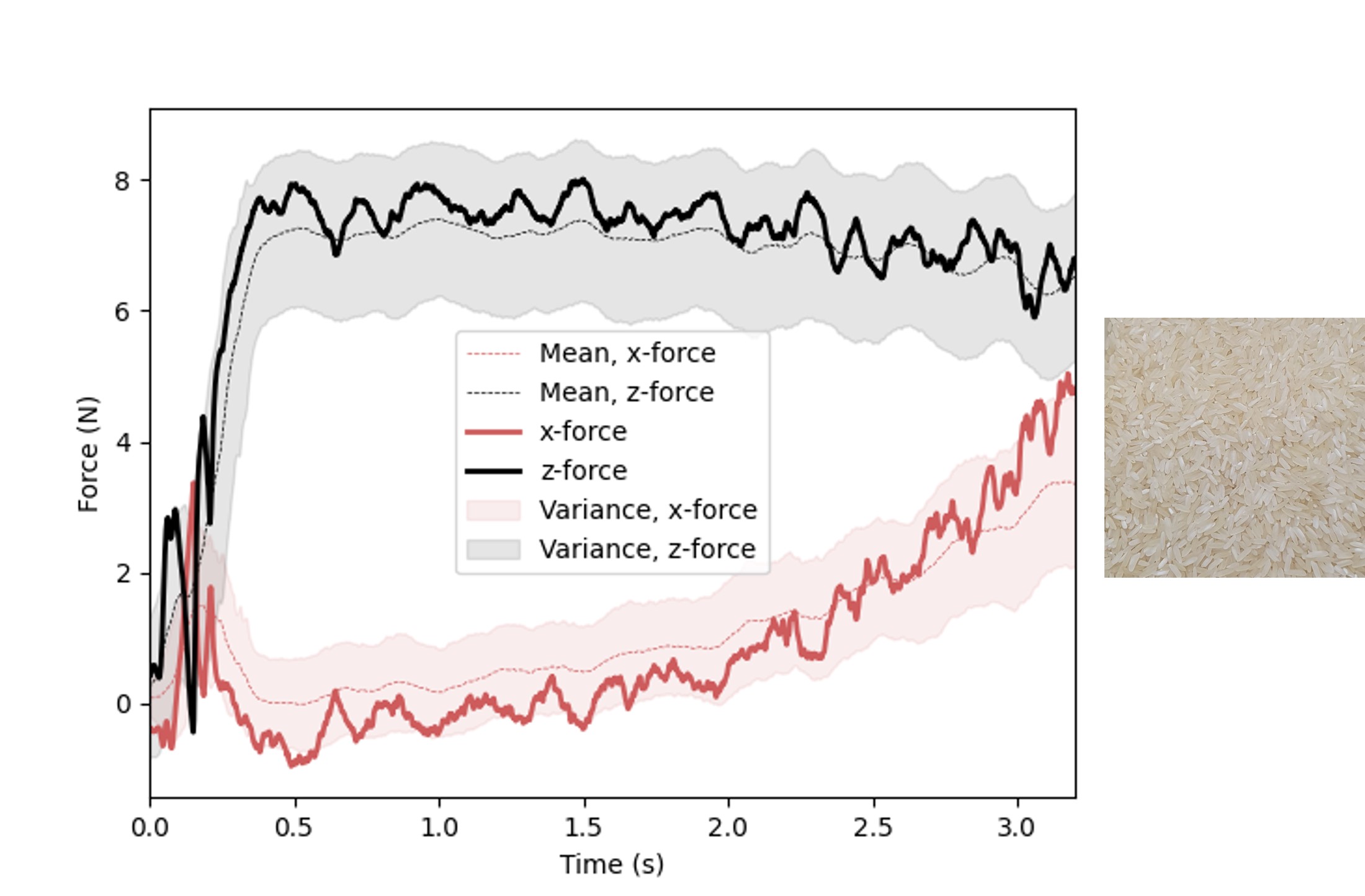}
        \captionsetup{justification=centering}
        \caption{Rice}
    \end{subfigure}
    \begin{subfigure}[t]{0.32\linewidth}
        \centering
        \includegraphics[width=\linewidth]{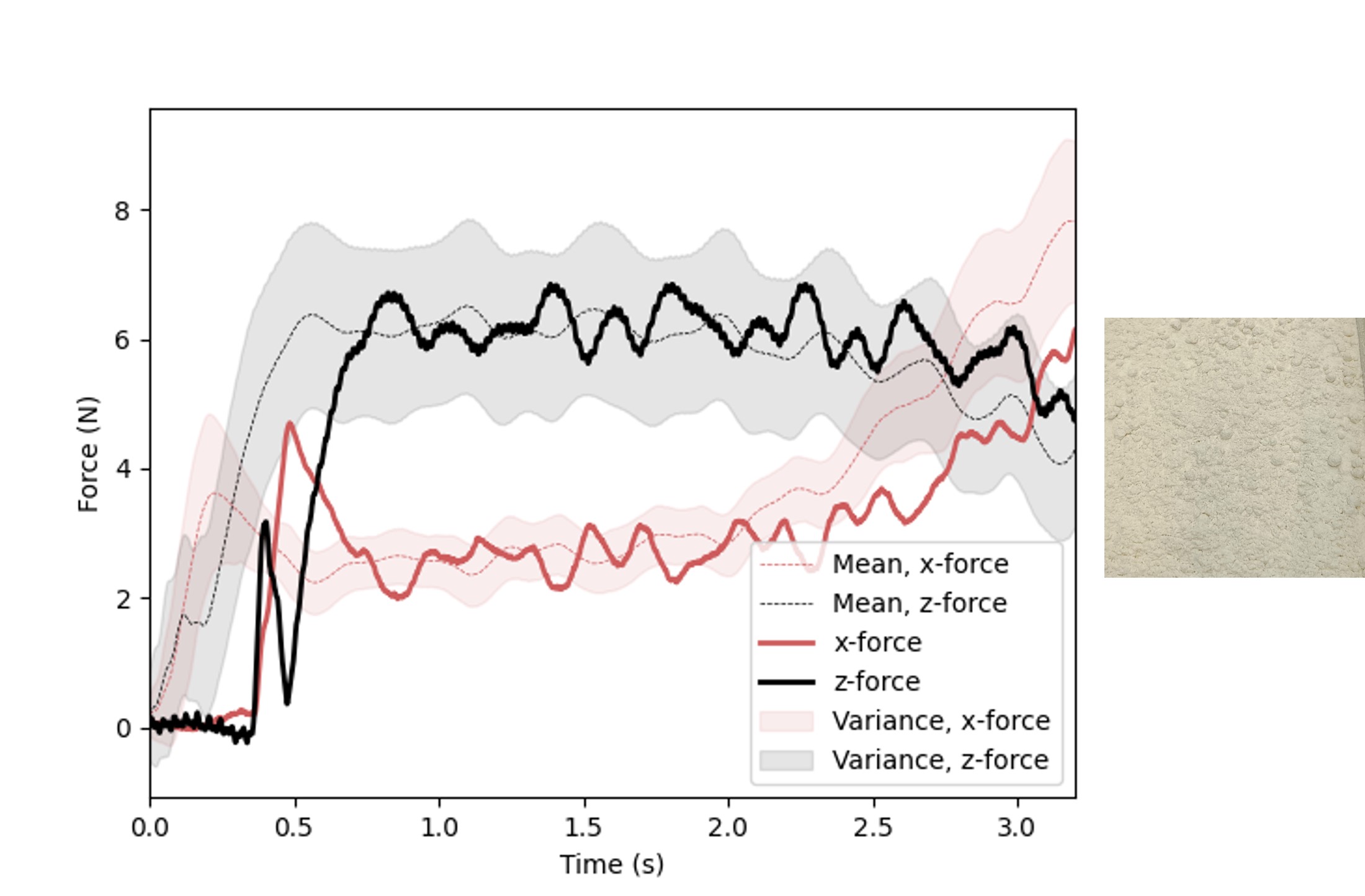}
        \captionsetup{justification=centering}
        \caption{Wheat flour}
    \end{subfigure}
    \begin{subfigure}[t]{0.32\linewidth}
        \centering
        \includegraphics[width=\linewidth]{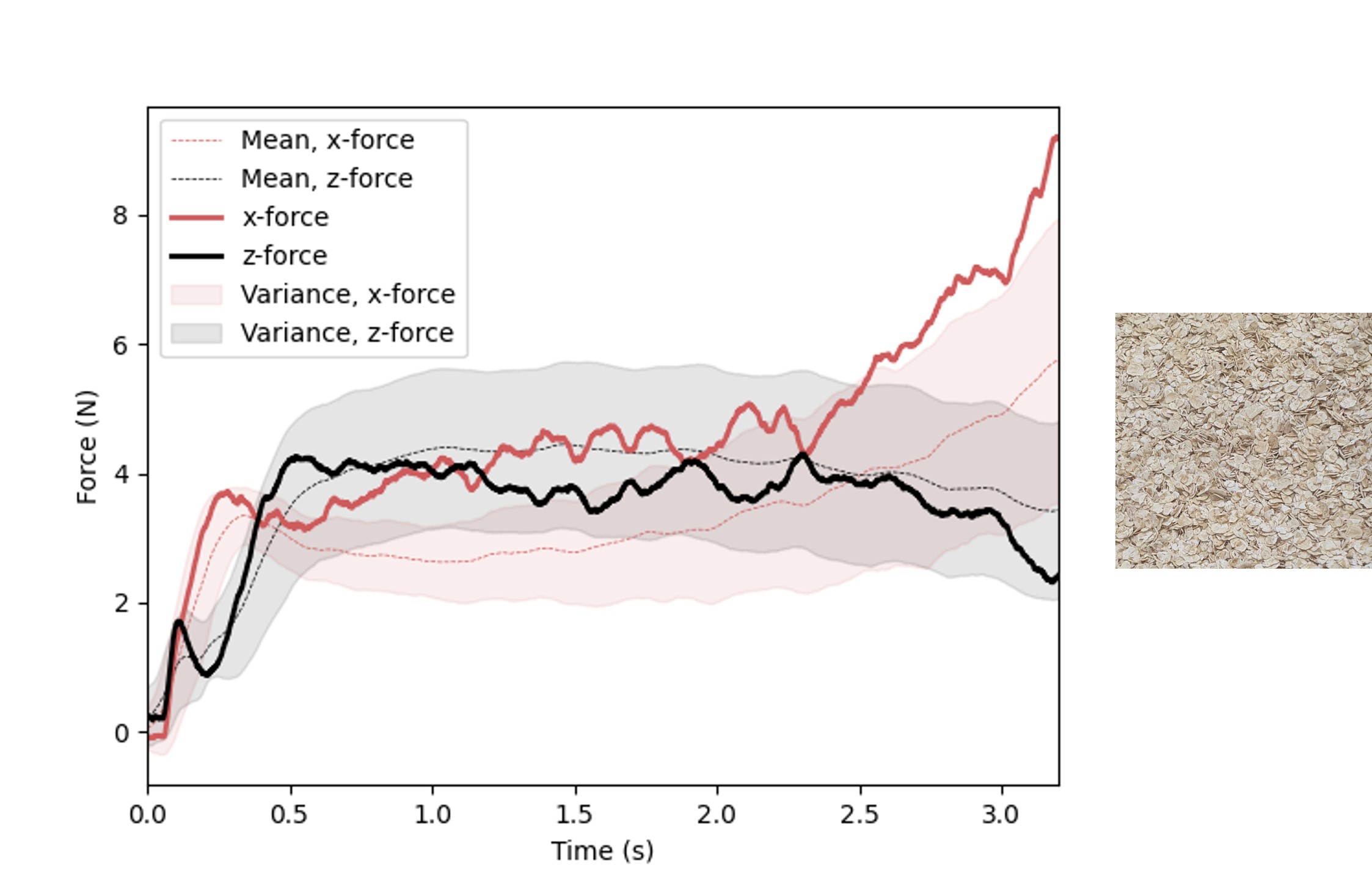}
        \captionsetup{justification=centering}
        \caption{Oat flakes}
    \end{subfigure}
    \begin{subfigure}[t]{0.32\linewidth}
        \centering
        \includegraphics[width=\linewidth]{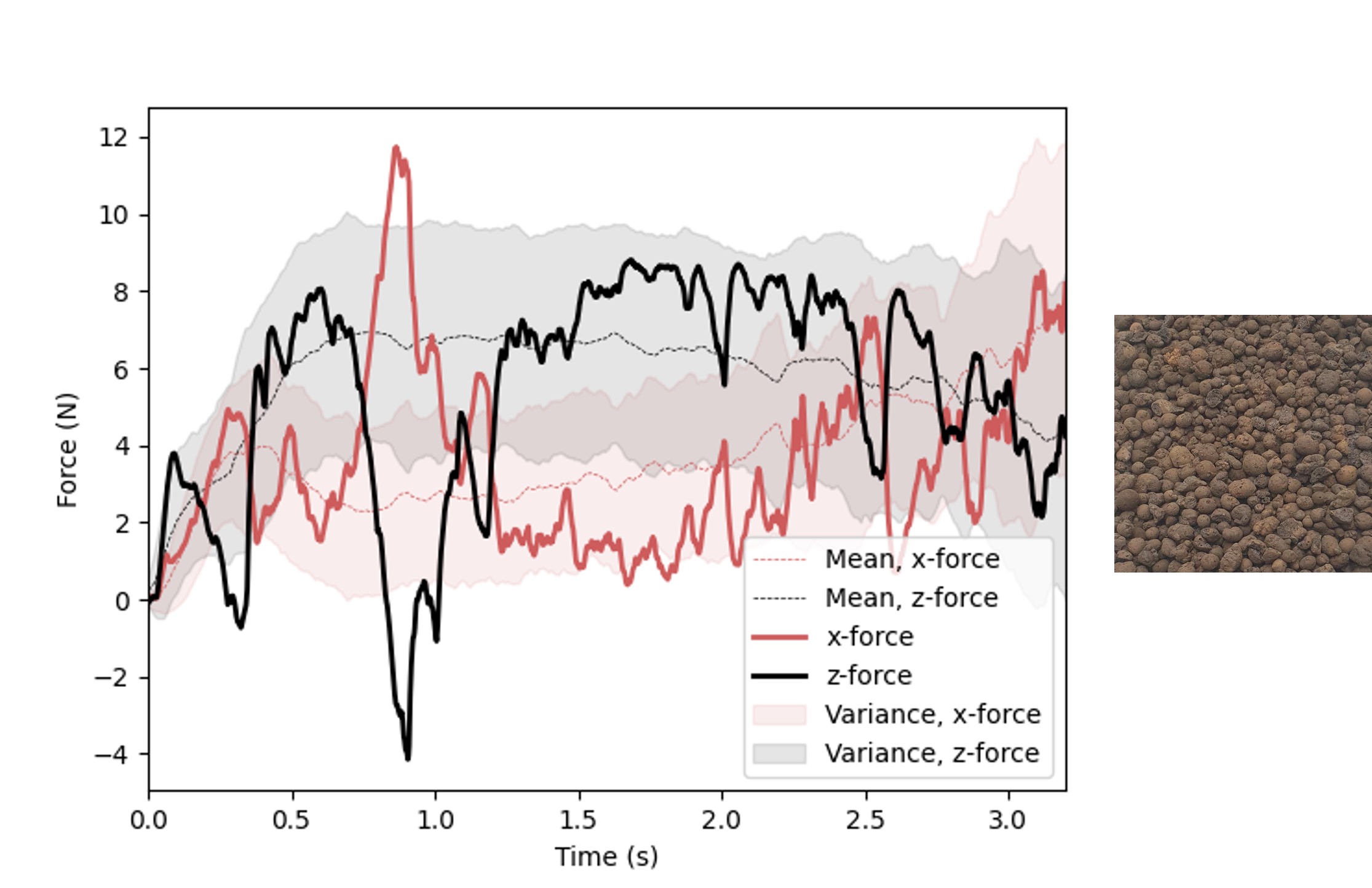}
        \captionsetup{justification=centering}
        \caption{Potting gravel}
    \end{subfigure}
    \begin{subfigure}[t]{0.32\linewidth}
        \centering
        \includegraphics[width=\linewidth]{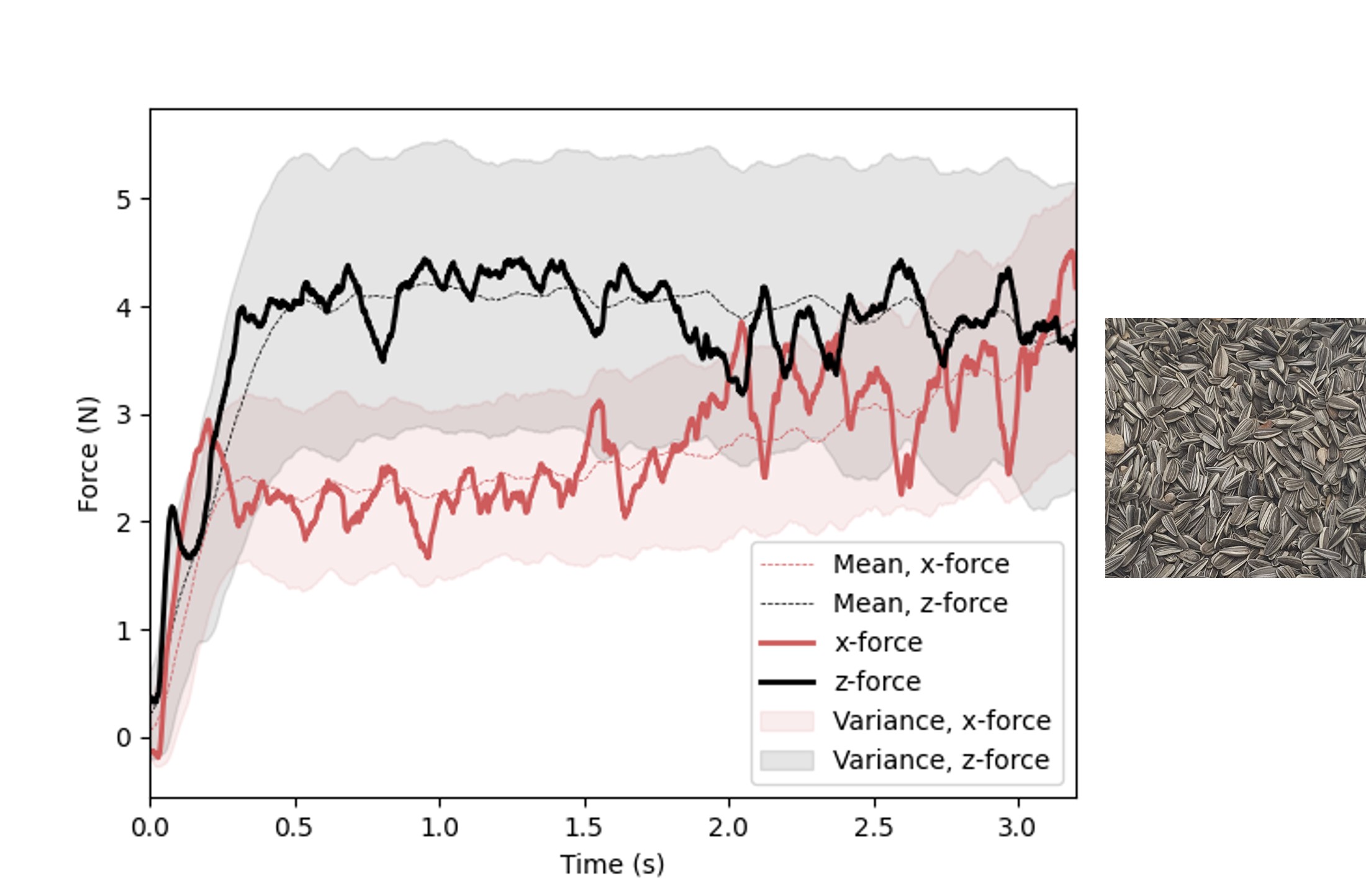}
        \captionsetup{justification=centering}
        \caption{Sunflower seeds}
    \end{subfigure}
    \begin{subfigure}[t]{0.32\linewidth}
        \centering
        \includegraphics[width=\linewidth]{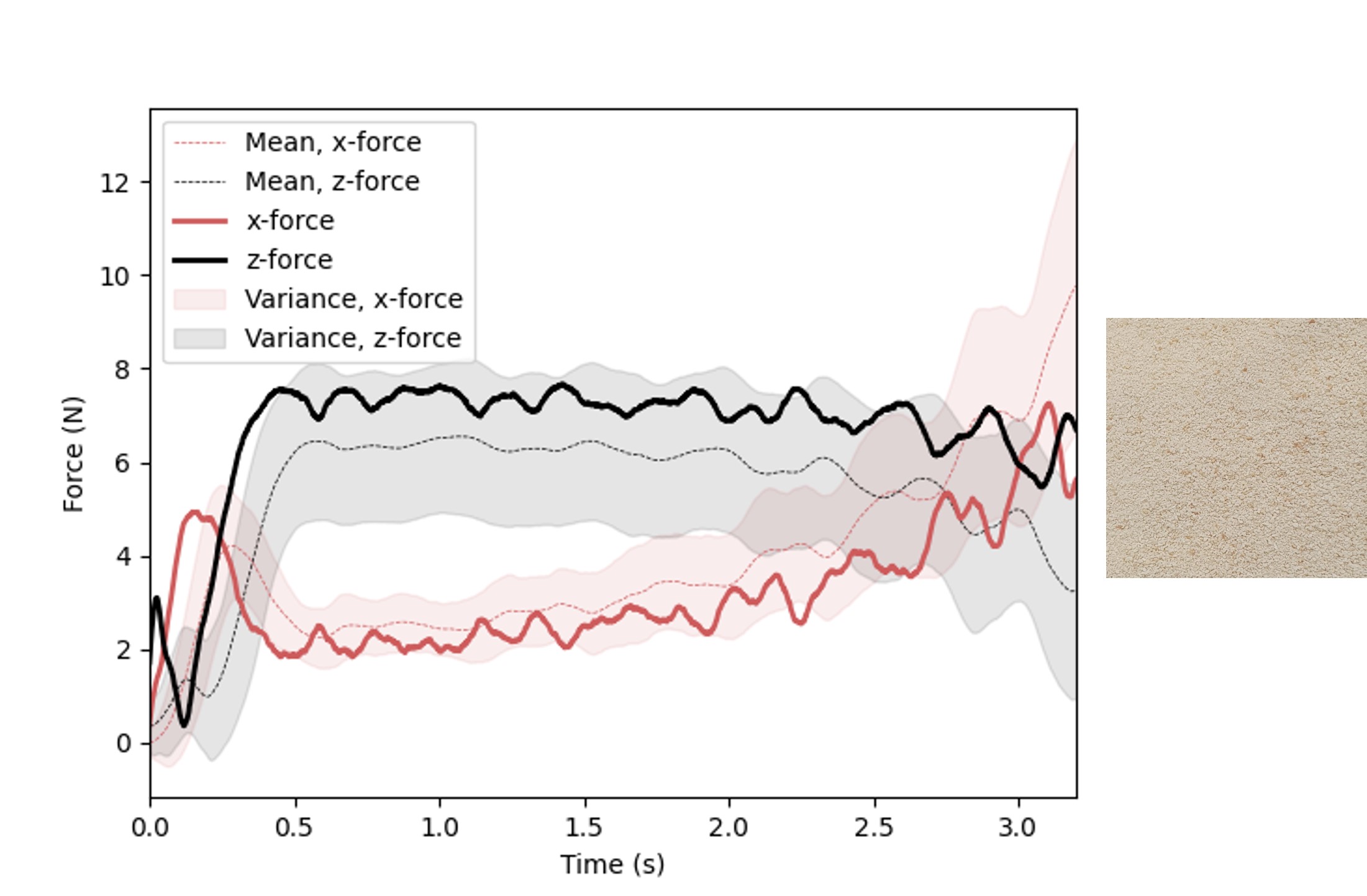}
        \captionsetup{justification=centering}
        \caption{Breadcrumbs}
    \end{subfigure}
\begin{subfigure}[t]{0.32\linewidth}
        \centering
        \includegraphics[width=\linewidth]{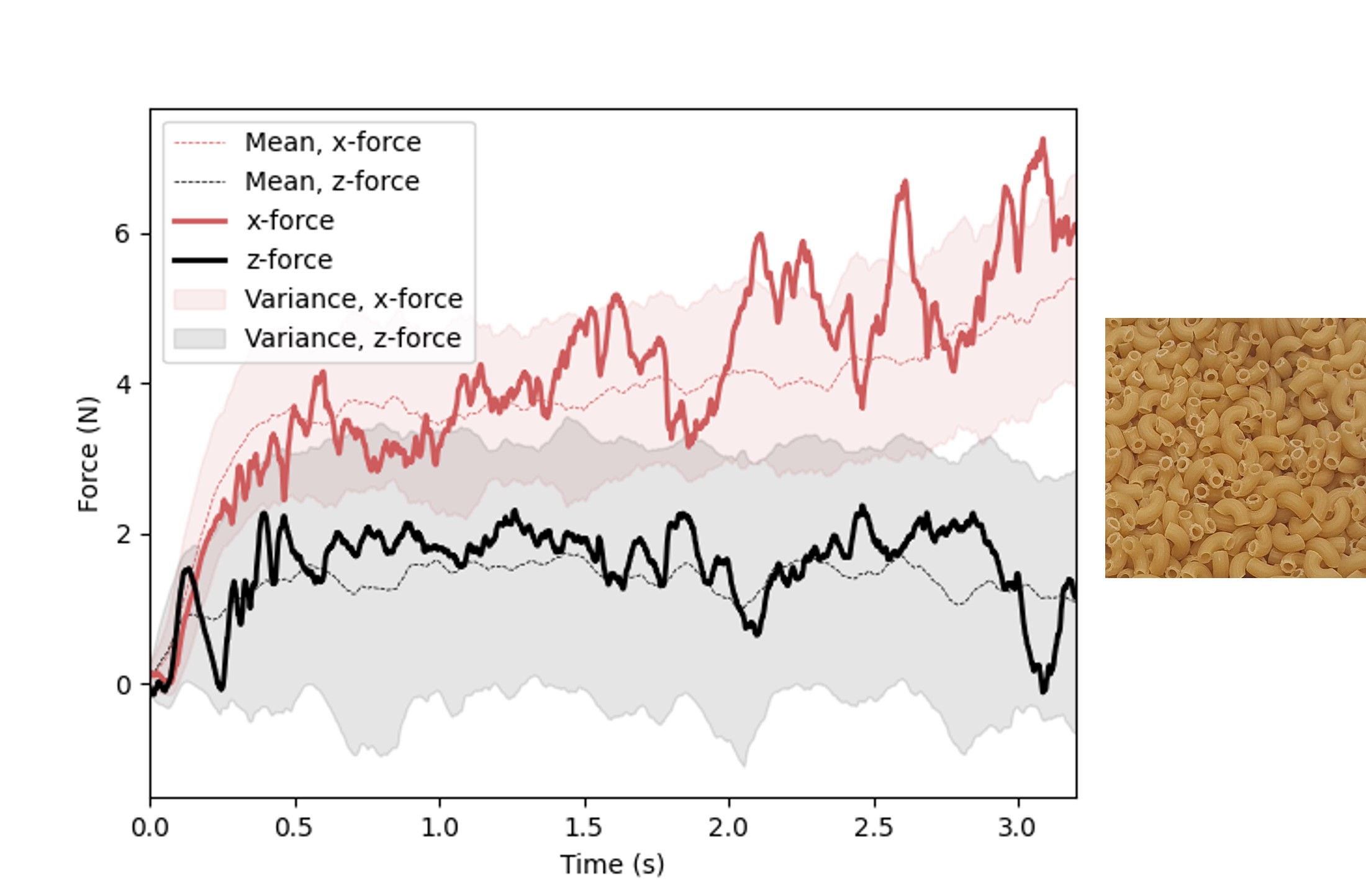}
        \captionsetup{justification=centering}
        \caption{Macaroni}
    \end{subfigure}
    \begin{subfigure}[t]{0.32\linewidth}
        \centering
        \includegraphics[width=\linewidth]{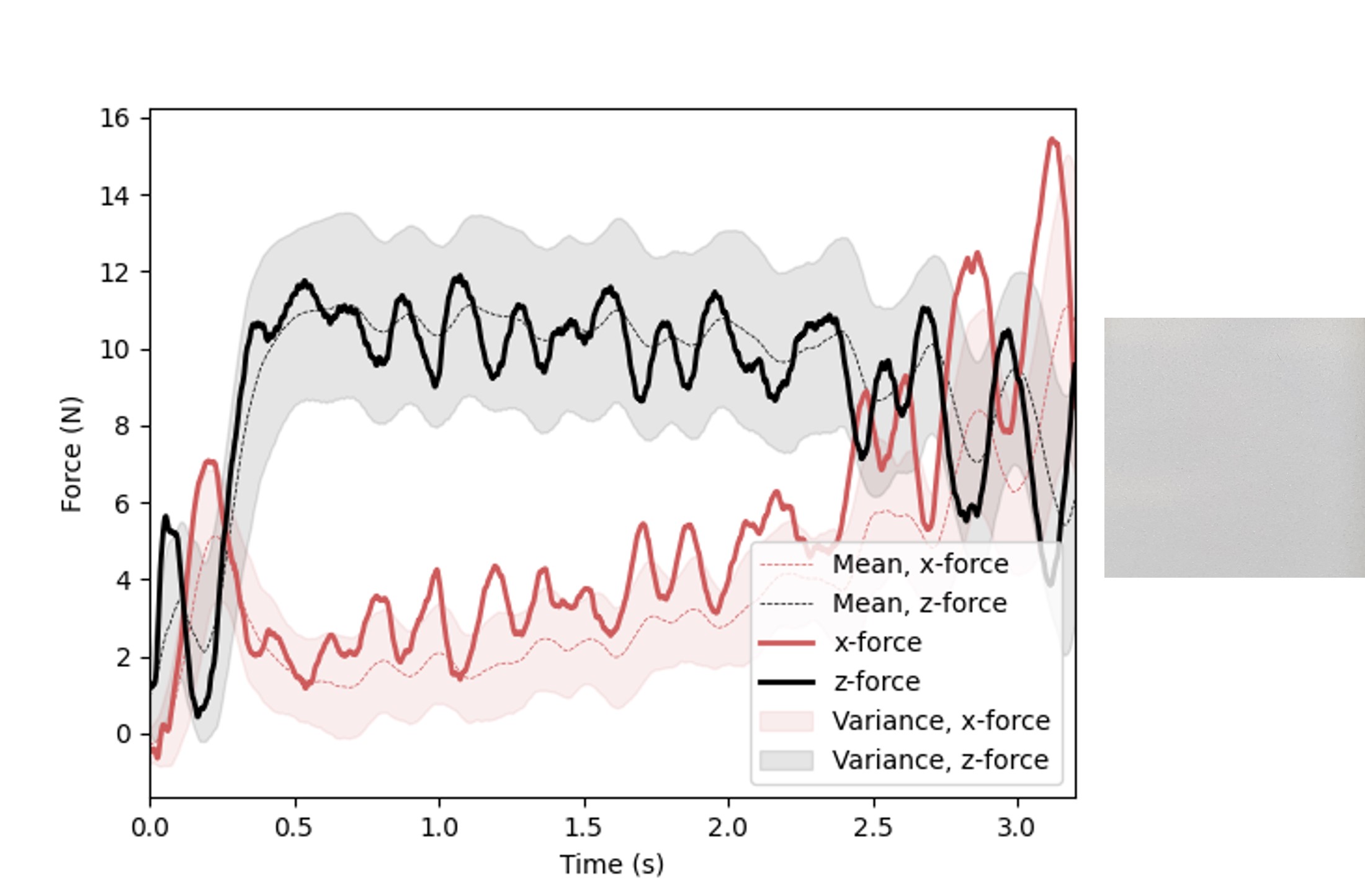}
        \captionsetup{justification=centering}
        \caption{Fine sugar}
    \end{subfigure}
    \begin{subfigure}[t]{0.32\linewidth}
        \centering
        \includegraphics[width=\linewidth]{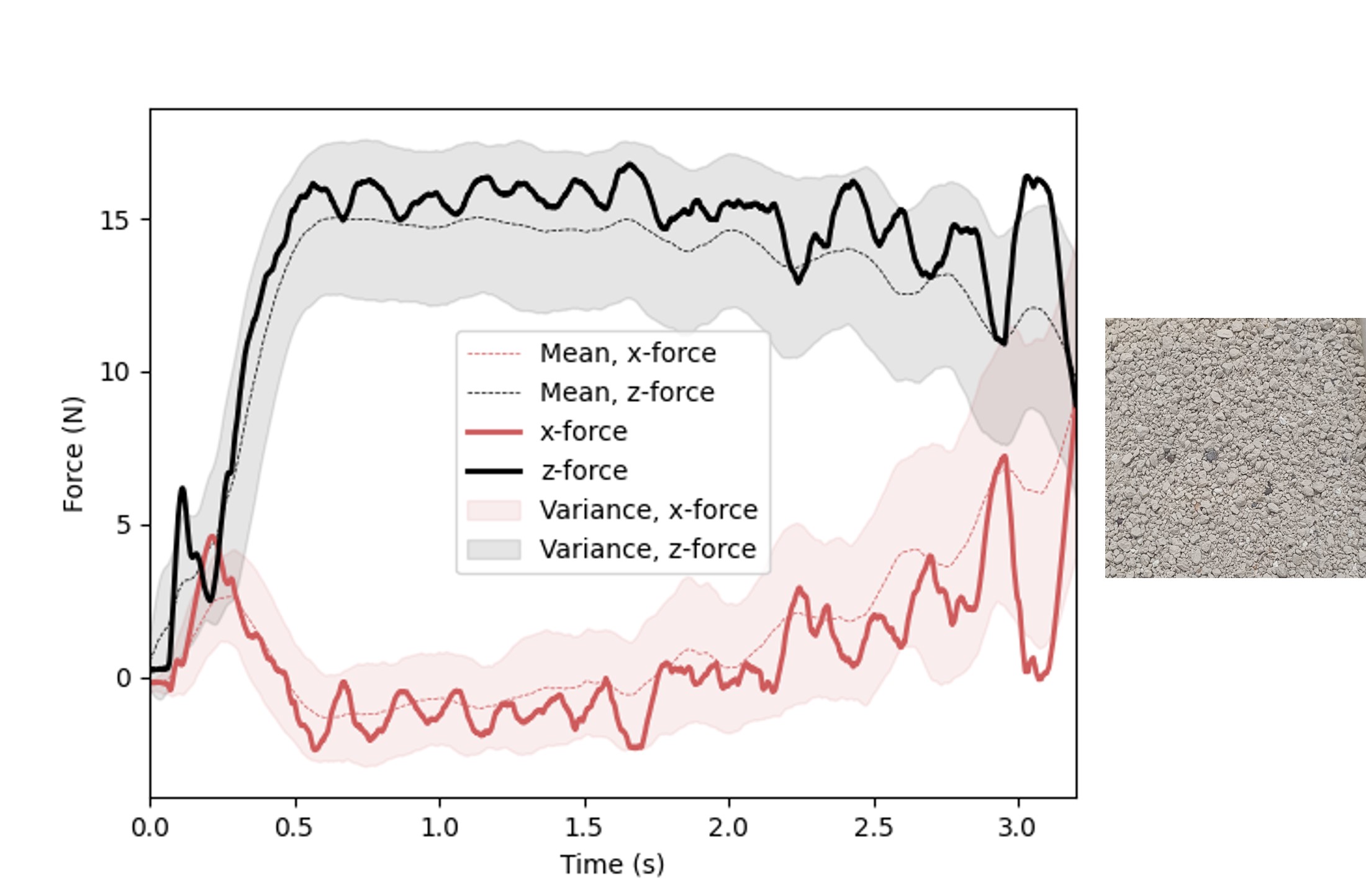}
        \captionsetup{justification=centering}
        \caption{Cat litter}
    \end{subfigure}
        \begin{subfigure}[t]{0.32\linewidth}
        \centering
        \includegraphics[width=\linewidth]{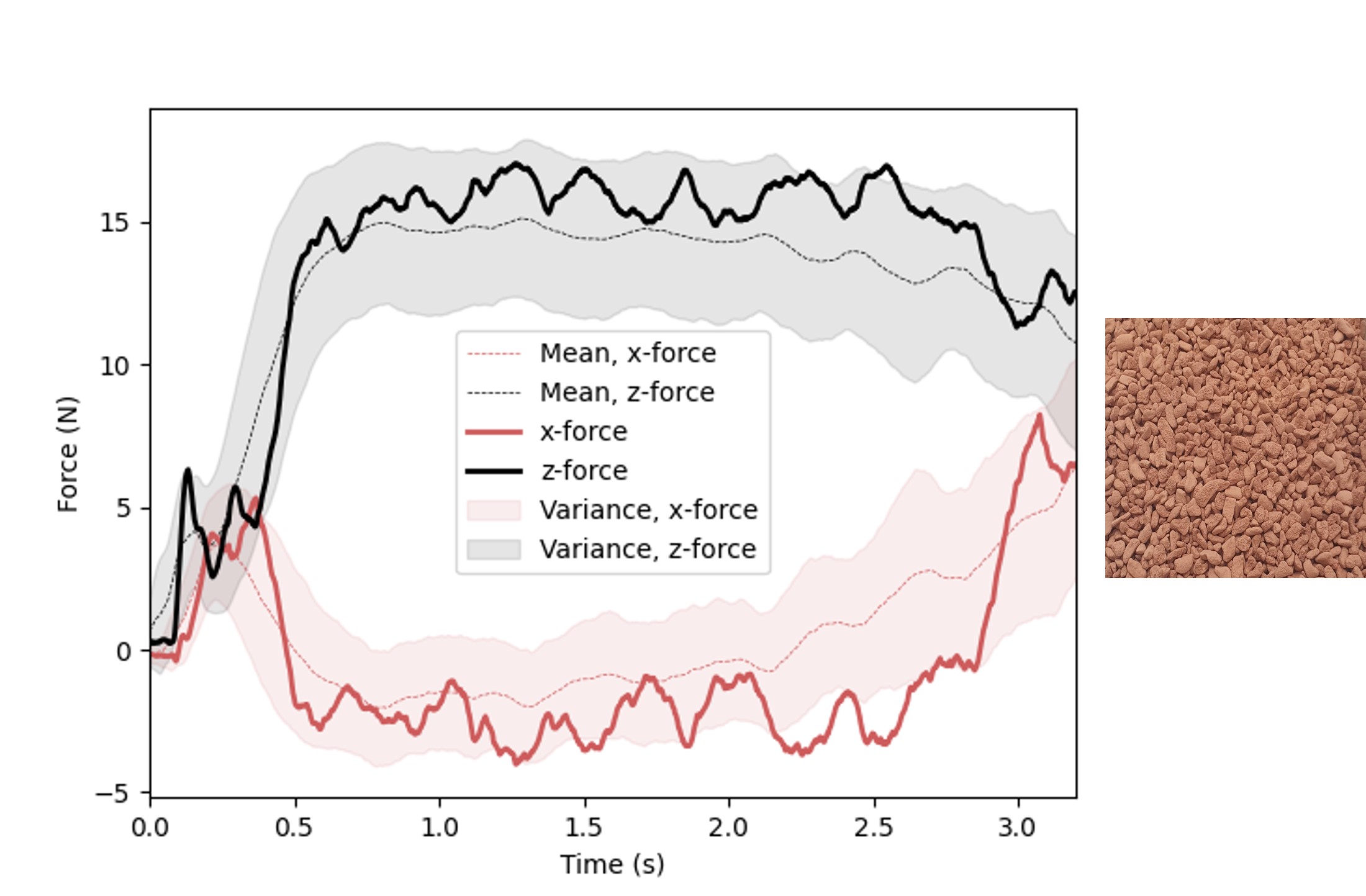}
        \captionsetup{justification=centering}
        \caption{Clay granules}
    \end{subfigure}
    
\caption{Materials and examples of their F/T measurements on the x- and z-axis, together with mean-variance plots (zoom in for a better view)}
\label{fig:data_kuvat}
\end{figure*}

To gain deeper insights into the nature of the granular materials and how they differ, we conducted a comprehensive qualitative data analysis. The analysis aimed to examine whether the materials have enough differences to be classified and what the features distinguishing them are. The most significant forces were recorded along the x-axis and z-axis, while other forces remained closer to zero and less meaningful. Interestingly, forces along the z-axis tend to somewhat mirror those of the x-axis. Figure~\ref{fig:data_kuvat} illustrates typical F/T measurements obtained from interactions with each material. This figure presents the forces exclusively along the x- and z-axis to facilitate the inclusion of mean-variance plots while maintaining clarity and avoiding visual clutter. It can be observed that there are differences in both the magnitude of the measurements and their high-frequency content. Furthermore, we can observe differences in the signal dynamics, that is, in the way the signals evolve from the initial samples toward the end of the signal. Although it is impossible to comprehensively analyze all materials separately within a limited space, in the following, we will discuss the distinctive features at a general level while highlighting the most interesting examples.

\subsubsection{Signal magnitude}
The most straightforward distinguishable factor between the different materials is the magnitude of the measurements, referred to as the signal magnitude. The signal magnitude indicates how much force is needed at each point in time during the interaction to maintain the constant end-effector velocity. Thus, the signal magnitude reflects information about the material's flowing resistance, which correlates with its density and friction properties. Figure \ref{fig:data_kuvat} highlights the differences between the materials. For instance, the average magnitude of high-density materials, like fine sugar, is higher than that of lower-density materials, like sunflower seeds. However, relying solely on signal magnitude appears inadequate for identification, especially as the number of materials increases. This is evidenced, for example, by the similarities in the average signal magnitude of materials such as oat flakes and breadcrumbs. We evaluate the importance of the signal magnitude information empirically in Section \ref{sec:exp_and_res} by comparing the classification results achieved with raw signals to the results with signals that are normalized between 0 and 1. We also compare the results when normalized and non-normalized signals are used together with the proposed HFMH feature space. 

\subsubsection{Frequency content}
Figure \ref{fig:data_kuvat} also highlights distinct frequency-related patterns exhibited by various materials. For example, rice yields low-amplitude, rapid zig-zag patterns, whereas wheat flour demonstrates smoother behavior. Fine sugar shows consistent and relatively smooth, high-amplitude wave patterns. Macaroni, on the other hand, produces measurements with irregular and sharp turns. As observed, materials with larger grain sizes or high friction, stemming from irregular grain shapes or surface properties, tend to produce sharper and more irregular patterns in the measurements. These properties determine how smoothly the robot end-effector can traverse the material. 

Although the patterns affect the signals' frequency domain representations, pure frequency domain signals seem to lack distinguishable patterns. We will also show in the section \ref{sec:exp_and_res} that classification results using pure frequency domain representations are sub-optimal. For this reason, our proposed method does not utilize the raw frequency information but instead uses a more compact presentation of the signal energy in high frequencies, as described in Section \ref{sec:method}.

\subsubsection{Signal dynamics}
Signal dynamics refer to the qualitative patterns of how the signals develop from the beginning to the end. The signals, in general, exhibit non-stationarity, with the majority showing an increasing trend on the x-axis (decreasing on the z-axis after the initial peak). For forces along the x-axis, a rising trend is expected since the end-effector pushes the material forward during movement, causing the material to compress between the end-effector and the back edge of the container. Thus, a higher force is needed towards the end of the motion trajectory to maintain a constant velocity.

Some of the materials also display unique dynamics. For example, rice produces an easily identifiable high peak on the x-axis at the beginning of the signal, after which the signal descends close to zero and starts to gradually increase again towards the end. Another interesting pattern can be seen for clay granules and other clay granule-like materials (cat litter and, to some extent, potting gravel). For clay granule-like materials, after an initial force increase on the x-axis, the readings take negative values. We associate this behavior with the dynamics of the material flow as the material moves away from the end-effector's path and returns to fill the groove.

In summary, our analysis demonstrates that direct interaction with various granular materials yields meaningful force measurements. Distinct characteristics emerge for different materials in both the time and frequency domain. The analysis in this section is also the basis for the design of the proposed feature space representation in this work. Furthermore, the provided analysis gives valuable insights for designing high-quality feature spaces that not only facilitate the classification of granular materials but also enable the estimation of more complex material properties, such as friction coefficients or flow dynamics, in the future.

%% file: sections/experiment.tex
\section{{Experiments and Results}}
\label{sec:exp_and_res}
\begin{table}[t]
    \centering
    \vspace{0.5cm}
    \ra{1.3}\tbs{15}
    \begin{tabular}{lc} 
        \toprule
              Features & Classification accuracy ($\pm$ SD)\\
        \midrule
        \textbf{Raw signal + HFMH}&\textbf{0.9705} ($\pm$ 0.0098) \\
        Raw signal + DFT & 0.7712 ($\pm$ 0.0426)\\
        Norm. raw signal + HFMH& 0.9303 ($\pm$ 0.0155)\\
        Norm. raw signal + DFT & 0.7686 ($\pm$ 0.0450) \\
        HFMH  &0.8705 ($\pm$ 0.0163)\\
        Raw signal &0.7614 ($\pm$ 0.0265)\\
        Raw hist. & 0.7072 ($\pm$ 0.0434) \\
        Norm. raw signal &0.8731 ($\pm$ 0.0251) \\
        DFT &0.7621 ($\pm$ 0.0371) \\
        PCA & 0.7527 ($\pm$ 0.0373) \\
        PCA + RICA & 0.7602 ($\pm$ 0.0288) \\
        \bottomrule
    \end{tabular}
    \caption{Classification accuracy for different feature spaces. The higher, the better.}
    \label{table:classification_accuracies1}
\end{table}

\begin{figure*}[t]
\vspace{1cm}
\centering
\begin{tikzpicture}[scale=1]
    \begin{axis}[
            colormap={bluewhite}{color=(white) rgb255=(90,96,191)},
            xlabel=\textbf{Predicted},
            xlabel style={yshift=0pt},
            ylabel=\textbf{Actual},
            ylabel style={yshift=0pt},
            xticklabels={Dry peas, Rice, Wheat flour, Clay granules, Oat flakes, Potting gravel, Sunflower seeds, Breadcrumbs, Macaroni, Fine sugar, Cat litter}, 
            xtick={0,...,10}, 
            xtick style={draw=none},
            yticklabels={Dry peas, Rice, Wheat flour, Clay granules, Oat flakes, Potting gravel, Sunflower seeds, Breadcrumbs, Macaroni, Fine sugar, Cat litter}, 
            ytick={0,...,10}, 
            ytick style={draw=none},
            enlargelimits=false,
            colorbar,
            xticklabel style={
              rotate=75,
            },
            nodes near coords={\pgfmathprintnumber[fixed,precision=2]\pgfplotspointmeta},
            nodes near coords style={
                yshift=-7pt
            },
        ]
        \addplot[
            matrix plot,
            mesh/cols=11, 
            point meta=explicit,draw=gray,
        ] table [meta=C] {
            x y C
            0 0 0.9750	
            1 0 0.0250	
            2 0 0
            3 0 0
            4 0 0
            5 0 0
            6 0 0
            7 0 0
            8 0 0
            9 0 0
            10 0 0
            
            0 1 0.0208
            1 1 0.9708
            2 1 0
            3 1 0
            4 1 0
            5 1 0
            6 1 0
            7 1 0
            8 1 0	 
            9 1 0
            10 1 0.0083
											
            0 2 0
            1 2 0
            2 2 1
            3 2 0
            4 2 0	
            5 2 0
            6 2 0
            7 2 0
            8 2 0 
            9 2 0
            10 2 0
            
            0 3 0
            1 3 0
            2 3 0
            3 3 0.9042	
            4 3 0
            5 3 0
            6 3 0
            7 3 0
            8 3 0.0083 
            9 3 0
            10 3 0.0875								
            
            0 4 0
            1 4 0
            2 4 0
            3 4 0
            4 4 0.9917
            5 4 0
            6 4 0
            7 4 0.0083
            8 4 0 
            9 4 0
            10 4 0

            0 5 0
            1 5 0
            2 5 0
            3 5 0
            4 5 0
            5 5 0.9917
            6 5 0
            7 5 0
            8 5 0.0083 
            9 5 0
            10 5 0

            0 6 0
            1 6 0
            2 6 0
            3 6 0
            4 6 0
            5 6 0
            6 6 1	
            7 6 0	
            8 6 0 
            9 6 0
            10 6 0
            
            0 7 0
            1 7 0
            2 7 0
            3 7 0
            4 7 0.0417
            5 7 0
            6 7 0
            7 7 0.9542 
            8 7 0 
            9 7 0
            10 7 0
            
            0 8 0.0167
            1 8 0	
            2 8 0
            3 8 0.0125	
            4 8 0
            5 8 0
            6 8 0
            7 8 0
            8 8 0.9708
            9 8 0
            10 8 0
            
            0 9 0
            1 9 0
            2 9 0
            3 9 0
            4 9 0
            5 9 0
            6 9 0
            7 9 0
            8 9 0 
            9 9 1
            10 9 0
            
            0 10 0
            1 10 0.0250
            2 10 0
            3 10 0.0333
            4 10 0
            5 10 0
            6 10 0.0250
            7 10 0 
            8 10 0 
            9 10 0
            10 10 0.9167
        }; 
    \end{axis}
\end{tikzpicture}
\caption{Confusion matrix for classification with raw + HFMH features on the test dataset.}
\label{table:confusion_1}
\end{figure*}

The goal of our experiments is to answer the following questions:
\begin{enumerate}
  \item What level of classification accuracy can be achieved by the proposed interactive perception framework powered with the ECOC-SVM classifier?
  \item How does the proposed feature space, raw signal + HFMH, perform compared to the baseline feature representations?
  \item How do the distinguishing signal characteristics, described in Section \ref{sec:data_analysis_B}, affect the classification accuracy? 
\end{enumerate}

We tested the proposed method by splitting the dataset into training and test sets randomly 20 times so that 50 samples out of each class are always used for training and 12 samples for testing. We compared the proposed feature space, the raw signal plus its HFMH, to alternative feature spaces to assess its performance. The alternative feature spaces were the plain raw signal, the histogram of the raw signal, the frequency-domain signal (Discrete Fourier Transform, DFT), and selected combinations thereof. Additional comparisons to established feature extraction methods, Principal Component Analysis (PCA) \cite{pearson_1901_1430636} and PCA + Reconstruction Independent Component Analysis (RICA) \cite{le2011ica} were made. Further, we evaluated the importance of the major distinguishing characteristics, identified in section \ref{sec:data_analysis_B}, by alternately eliminating the effect of each. The average classification accuracy of each feature combination is reported in Table \ref{table:classification_accuracies1}.

The results show that ECOC-SVM with raw + HFMH features reaches over 97 \% classification accuracy on average. The second-best feature space is the normalized version of the same feature space. These feature spaces outperform the next ones, normalized signal, and plain HFMH by a margin of 6-10 \% percentage points.

Considering the established feature extractors, PCA did not yield statistically different results than the raw signal ($p < 0.2967$\footnote{To test the statistical significance of the differences, we performed a pairwise Mann-Whitney U-test. The null hypothesis is that the samples from two distributions have equal mean. The null hypothesis is rejected if $p$ is smaller than a certain threshold, for example, $p < 0.05$ for a 5 \% significance level.}). Neither did the results with PCA + RICA differ significantly from those with the raw signal ($p < 0.7968$). Plain PCA did not differ significantly from combined PCA and RICA either ($p < 0.5146$). The best number of features, 100 after PCA and 80 after RICA, was found experimentally.

The accuracy difference between the raw signal and the normalized raw signal provides insights into the importance of signal magnitude. Interestingly, the normalization of the signal, despite losing information, clearly increases the classification accuracy ($p < 5.8351\cdot 10^{-8}$). However, when combining the raw signal with HFMH, the normalization leads to decreased classification accuracy ($p < 1.1161\cdot 10^{-7}$). Even though the results seem conflicting at first glance, they imply that the signal magnitude carries valuable information about the material. Still, the ECOC-SVM classifier finds better decision boundaries for raw signals when the data is normalized. However, this benefit is lost when the raw signal is combined with other features. 

Similarly, we evaluated the significance of signal dynamics by comparing the results with the raw signal against the results with the histogram of the raw signal. The histogram representation diminishes all the information related to the signal dynamics but preserves magnitude information. Table~\ref{table:classification_accuracies1} shows that the histogram representation decreases classification accuracy by approximately 6 percentage points. The difference is not substantial but significant, implying that the signal dynamics is important information for accurate classification. To evaluate whether the performance decrease is, in fact, caused by the loss of dynamics information and not simply by more compact feature space, we also tried to test the same effect by using randomly permuted signals as the model input. However, we achieved only approximately 20 \% classification accuracy, likely due to too complex decision boundaries with respect to the available data.  

The results in this section show that the proposed interactive perception framework can recognize a range of granular materials with exceptional precision. Despite high classification accuracy, modest confusion was observed regarding gardening clay granules (simply called clay granules in Figure \ref{table:confusion_1}) and cat litter. Considering the similarity of the two, both being a type of clay granule, some confusion is expected. For other classes, the model reaches over 98 \% classification accuracy on average. The full confusion matrix is depicted in Figure ~\ref{table:confusion_1}. The proposed feature space and its normalized variation were the best feature spaces by a significant margin, the first one outperforming the third best feature space by approximately 10 percentage points. Furthermore, the results empirically validate our findings in the qualitative analysis in Section \ref{sec:data_analysis_B}. The results show that the best feature space combines all three main pieces of information identified in Section \ref{sec:data_analysis_B}. We also isolated the effect of each information category and analyzed their significance individually. 

%

%% file: sections/limitations_future.tex
\section{Limitations and Future Work}

While we report near-perfect classification accuracy in our experiments, the proposed method is not without its limitations. First of all, the method, as it is, assumes a fixed end-effector geometry, and the generalization ability to different end-effector shapes remains unknown. Furthermore, the current work does not address scenarios in which the material surface is too low for the end-effector to be fully submerged. Future work should address these limitations by incorporating varying end-effector geometries and container fill levels during data collection.

Some of the classification errors could be avoided with simple sensor fusion using a regular RGB-camera or even an RGBD-camera. For example, cat litter and gardening clay granules are easily confused with force sensing alone. However, the color difference of the materials would make them trivially classifiable with an RGB-camera. Also, sensor fusion with GelSight-type visuo-tactile sensors could be consider for better classification across a wider range of materials. 

Finally, many use cases would benefit more from the identification of material properties, such as particle size (distribution), shear strength, and friction parameters. The parameter estimation would allow better generalizability to new materials, and could also be more suitable for the optimization of downstream manipulation tasks, where knowledge of material properties is often more important than its class label. 

%% file: sections/conclusion.tex
\section{Conclusions}
\label{sec:conclusions}
We presented a new interactive perception framework that identifies granular materials through direct interaction. 
Furthermore, we described a novel, open dataset of F/T measurements of various granular materials, accompanied by a comprehensive qualitative analysis for further advancement in the area. The results demonstrate that the proposed framework can identify materials with high precision. 

While this work focused only on material identification, it also opens a path towards using F/T sensing to estimate granular material parameters, such as friction coefficients. This capability would be beneficial in tasks such as autonomous earthmoving. Also, the ability to detect subtler differences would open possibilities in applications such as cooking. This opens an interesting avenue for studying comprehensive estimation of the material properties which would require combining F/T sensing with other modalities such as RGB cameras or visuotactile sensors.
